\documentclass[11pt,a4paper]{article}
\usepackage[a4paper,margin=2cm,head=15pt]{geometry}
\usepackage{mathtools,amsmath,amsthm,amsfonts,amssymb,latexsym}
\usepackage{hyperref}
\usepackage{graphicx}
\usepackage{subcaption}
\usepackage{array}

\usepackage[square,sort,comma,numbers]{natbib}
\usepackage[utf8]{inputenc}
\usepackage{algorithm}
\usepackage{algpseudocode}
\usepackage[T1]{fontenc}
\usepackage{lmodern}
\usepackage{indentfirst}
\usepackage{xcolor}
\usepackage{multicol}
\usepackage{titling}
\usepackage{fancyhdr}
\usepackage{wrapfig}
\usepackage{multirow}

\usepackage{breqn}

\newcommand{\theyear}{2020}
\newcommand{\themonth}{April}

\newcommand{\thejournallong}{INTERNATIONAL JOURNAL OF COMPUTERS COMMUNICATIONS \& CONTROL}
\newcommand{\theonlineissn}{1841-9844}
\newcommand{\theissn}{1841-9836}
\newcommand{\thevolume}{15}
\newcommand{\theissue}{2}
\newcommand{\thedoi}{doi.org/10.15837/ijccc.2020.2.3862}
\newcommand{\theartnum}{3862}

\fancypagestyle{firstpage}
{
    \fancyhf{}

}

\usepackage{fancyhdr}

\begin{document}

\thispagestyle{firstpage}
\vspace*{\dimexpr-\headheight-\headsep}%
         \hrule\vspace{0.1cm}\hrule\vspace{0.2cm}
            \thejournallong\\
            Online ISSN \theonlineissn, ISSN-L \theissn, Volume: \thevolume, Issue: \theissue, Month: \themonth, Year: \theyear\\
            Article Number: \theartnum, \thedoi\\
        \hrule\vspace{0.1cm}\hrule\vspace{0.2cm}
        \includegraphics[width=2.8cm]{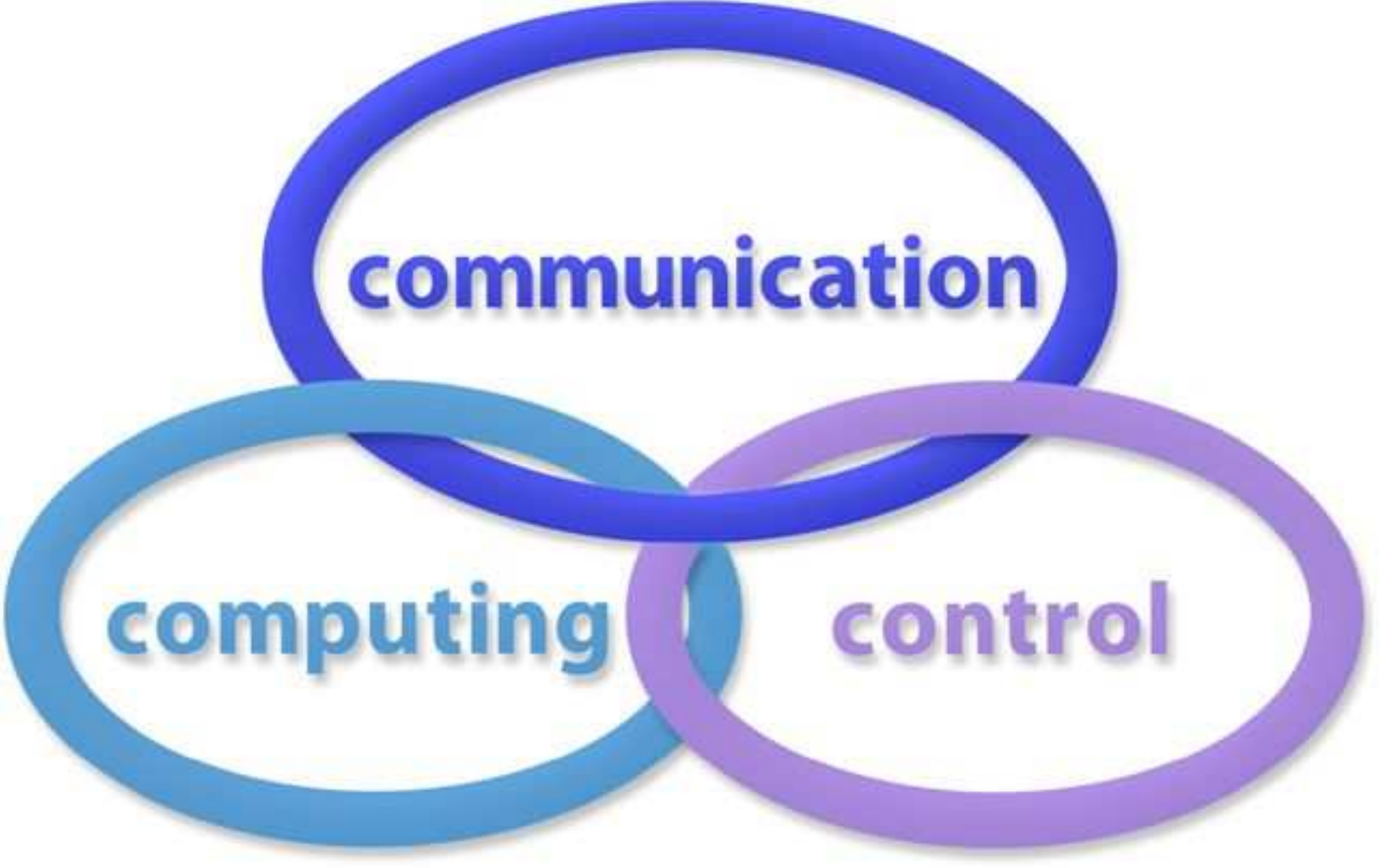}
            \hspace{3.2cm}
        \includegraphics[width=3.4cm]{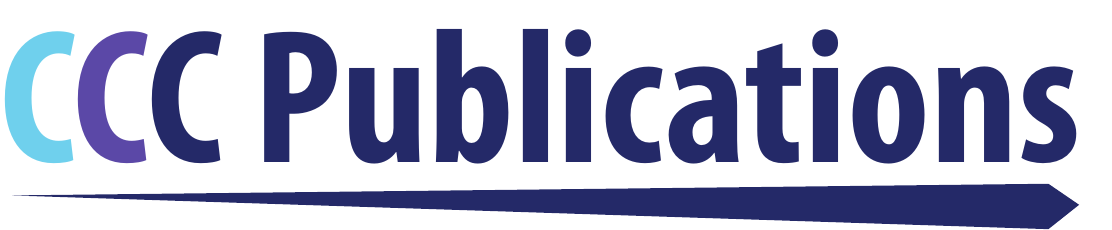}
            \hspace{3.2cm}
        \includegraphics[width=2.7cm]{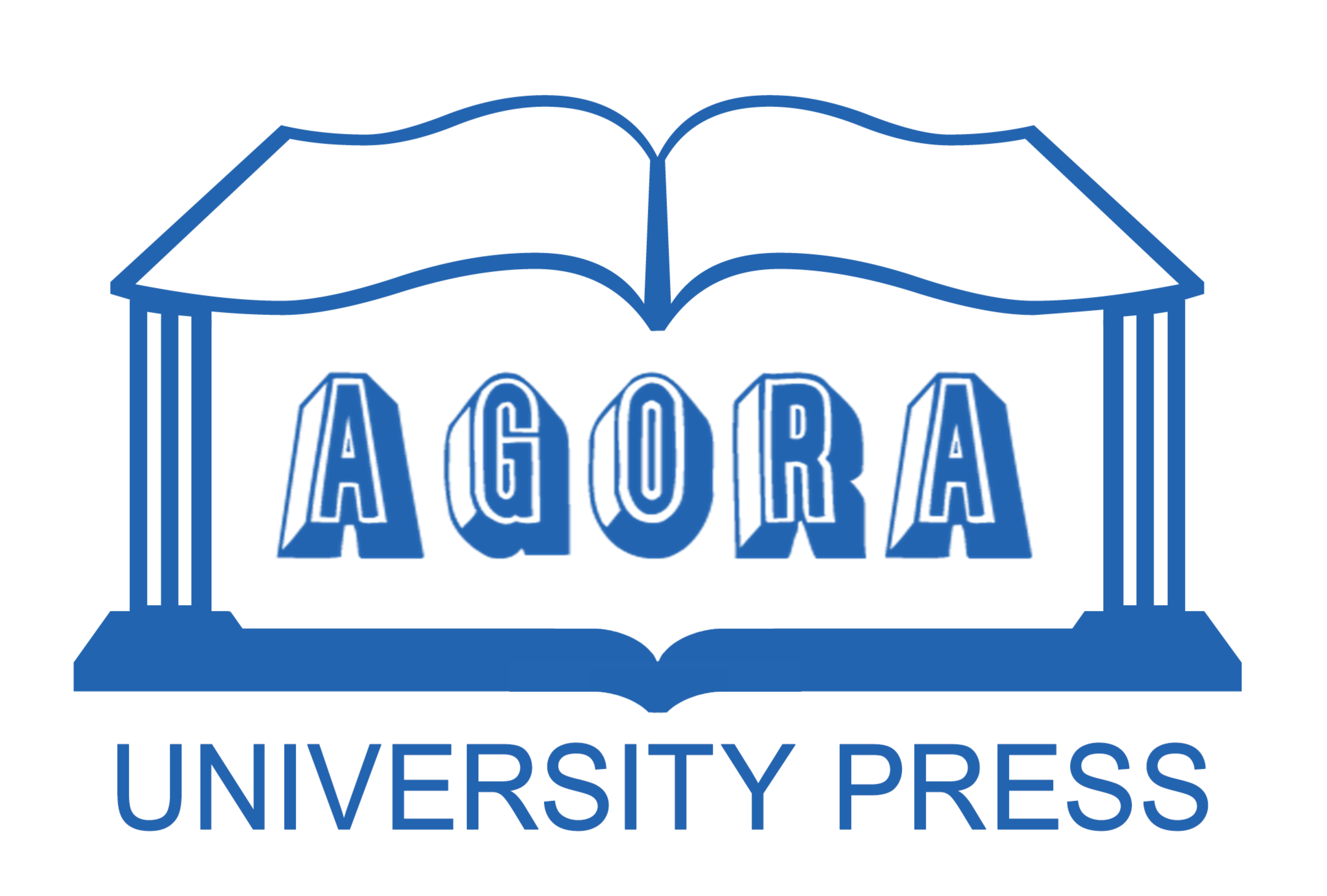}\\
        \hrule\

\title{Pose Manipulation with Identity Preservation}

\author{A. T. Ardelean, L. M. Sasu}
\date{}
{\let\newpage\relax 
\maketitle}
\thispagestyle{firstpage} 

{\small

\begin{flushleft}
\textbf{Andrei-Timotei Ardelean*}\\
Transilvania University of Brasov\\
500036 Brasov, B-dul Eroilor, 29, Romania\\
{*}Corresponding author: andrei.ardelean@student.unitbv.ro
\end{flushleft}

\begin{flushleft}
\textbf{Lucian Mircea Sasu}\\

1. Transilvania University of Brasov\\
500036 Brasov, B-dul Eroilor, 29, Romania\\
lmsasu@unitbv.ro\\
2. Xperi Corporation\\
500152 Brasov, Turnului, 5, Romania\\
lucian.sasu@xperi.com\\
\end{flushleft}

\begin{abstract}
This paper describes a new model which generates images in novel poses e.g. by altering face expression and orientation, from just a few instances of a human subject. Unlike previous approaches which require large datasets of a specific person for training, our approach may start from a scarce set of images, even from a single image. To this end, we introduce Character Adaptive Identity Normalization GAN (CainGAN) which uses spatial characteristic features extracted by an embedder and combined across source images. The identity information is propagated throughout the network by applying conditional normalization. After extensive adversarial training, CainGAN receives figures of faces from a certain individual and produces new ones while preserving the person's identity. Experimental results show that the quality of generated images scales with the size of the input set used during inference. Furthermore, quantitative measurements indicate that CainGAN performs better compared to other methods when training data is limited.

\end{abstract}
}

\section{Introduction}
Developing a way to easily manipulate face expression and head pose of an individual has been the focus of many research groups in the last decade. The baseline solution for this task involves creating a 3D model and animate it accordingly. However, building a photo-realistic head model using standard techniques can take substantial amount of time for a human artist. Many industries could benefit from optimizing this process such as cinematic, advertising and video games; furthermore, it has potential for image enhancement and editing software. 

A traditional automatic method for face manipulation is based on 3DMM fitting \cite{Blanz:1999:MMS:311535.311556}. Parameters can be estimated from a single image and then changed to obtain different expressions. This method is not sufficient by itself to work with hidden regions, e.g. teeth and closed eyes \cite{yuan2019face}.

Other approaches rely on warping from one or more source images to the desired pose to generate guided head images \cite{Wiles18}. A drawback of this solution is the limited amount of variation between the source and target pose it can manage without great loss of quality \cite{zakharov2019fewshot}.

New approaches for face image generation have been brought by recent advances in generative adversarial networks (GANs) \cite{NIPS2014_5423}. Seminal papers in this area \cite{karras2018progressive,tero2018style} show that one can generate high resolution realistic figures of human faces using GANs.

Our contribution is a new model, Character Adaptive Identity Normalization GAN (CainGAN), that receives figures of faces from a certain individual and produces new ones while preserving the person's identity. CainGAN generates images in novel poses starting from a small set of source pictures with the individual, i.e. a few-shot setting, without any fine-tuning as found in \cite{finn2017model,zakharov2019fewshot}.

We conducted experiments to compare images generated by CainGAN,  with alternative systems using image-to-image translation \cite{pix2pix2017,wang2018pix2pixHD} and conditioning based on Adaptive Instance Normalization \cite{huang2017adain} using computed embeddings \cite{zakharov2019fewshot}. By performing quantitative measurements on the self-reenactment task we show that our model is able to achieve state-of-the-art results using less data compared to other methods and without fine-tuning.  

The rest of the paper is structured as follows: In section \ref{sec:related} we make a literature review; the subsequent section describes CainGAN in detail; section \ref{sec:experiments} contains a comparison between different methods and an ablation study. Eventually, we summarize our contributions in section \ref{sec:conclusion}. Code for the implementation of CainGAN is available at \href{https://github.com/TArdelean/CainGAN}{https://github.com/TArdelean/CainGAN}

\section{Related Work}
\label{sec:related}
A number of works on face generation with preservation of identity focus on the talking face task, i.e. the area of interest is the mouth region with motion driven by either audio sources \cite{chen2018lip,XuSoundTV,suwajanakorn2017obama,zhou2019talking} or video to be imitated \cite{ijcai2019-129,zhou2019talking}. These methods cannot be easily extended to synthesize full head images that require handling more variation between poses or hidden elements in the source images. While it is possible to replace the face from an existing head footage, by using a face modeling approach as in \cite{thies2018face2face}, the result of pose manipulation would be limited to face expression.

Providing conditional information to several layers of the generator has been widely used to prevent input constraints from vanishing. Good results were obtained especially by modulating activations using AdaINs \cite{chen2018on,huang2017adain,tero2018style} and SPADE \cite{park2019SPADE} that employs spatial denormalization to introduce semantic map constraints.

Our model is based on a conditional GAN framework, i.e. instead of generating starting from noise as done traditionally \cite{NIPS2014_5423}, the input of the generative network can take different forms including images \cite{pix2pix2017,park2019SPADE} as done in this work. The use of multiple discriminators to stabilize GAN training has been recently studied in several works \cite{carneiro2019,Durugkar2016GenerativeMN,NIPS2017_6860}. Specifically our model uses two discriminators with different objectives.

\section{The CainGAN model}
In this section we present the architecture of the proposed model, followed by the training algorithm. Eventually we give some implementation details.
\subsection{The Architecture}
The goal is to train a generative model that is able to synthesize new images starting from $K$ existing source pictures with the same person. Let $x_i$ denote the $i$-th image of an image sequence $\mathbf{x}$; we uniformly sample $K+1$ distinct images from $\mathbf{x}$. The model receives as input $x_{i_1}, x_{i_2}, ..., x_{i_K}$ along with their corresponding landmark images \cite{bulat2017far} $L(x_{i_1}), L(x_{i_2}), ..., L(x_{i_K})$ and target landmark $L(x_{i_{K+1}})$, then generates a new image $\hat{x}_t$ that must follow target landmark and preserve identity of the person from the $K$ input images. The generated image is expected to be similar to the ground truth image $x_t := x_{i_{K+1}}$.

\begin{figure}[!t]
    \centering
    \includegraphics[width=\linewidth]{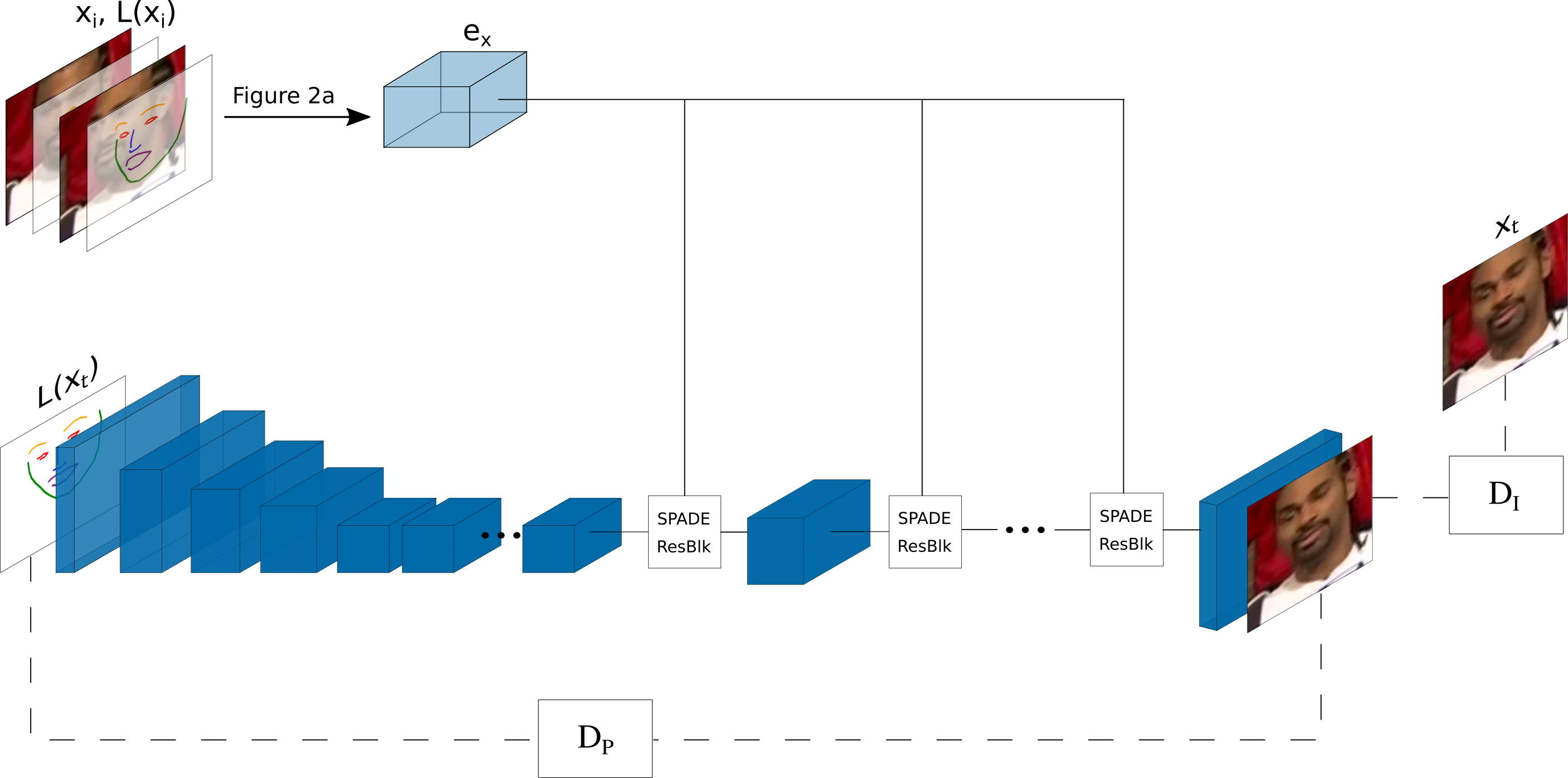}
    \caption{Full architecture and model workflow. Top side depicts the embedding which is used at each upsampling step to spatially modulate activations. The generator (shown in the bottom side of the figure) starts from the target landmarks $L(x_t)$ to synthesize a new image $\hat{x}_t$. $D_I$ and $D_P$ represent the identity and pose discriminators, respectively.}
    \label{fig:full_arhi}
\end{figure}

We propose CainGAN, a model that consists of 4 networks (Figure \ref{fig:full_arhi}) which we will describe in the following.  

\begin{figure}[!t]
\centering
\begin{subfigure}{0.57\textwidth}
    \includegraphics[height=12em]{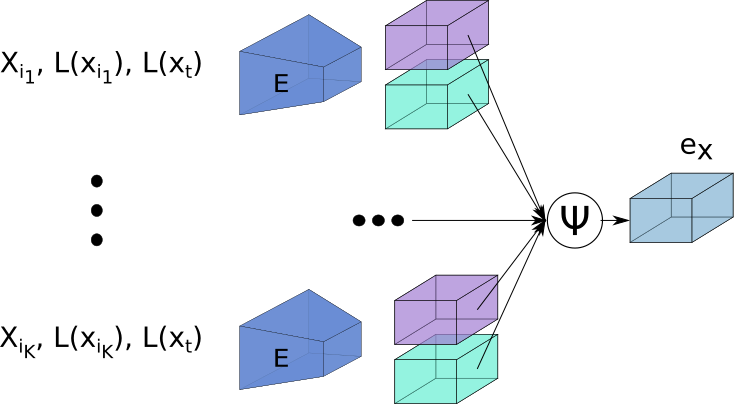}
    \caption{Combining embeddings based on responsibility, as in equation \eqref{eq:1}. The weights of the Targeted Embedder (E) are the same for each of the $K$ inputs. The Targeted Embedder is detalied in Figure \ref{fig:temb}.}
    \label{fig:comb_emb}
\end{subfigure}\hspace{0.04\textwidth}
\begin{subfigure}{0.37\textwidth}
    \includegraphics[height=12em]{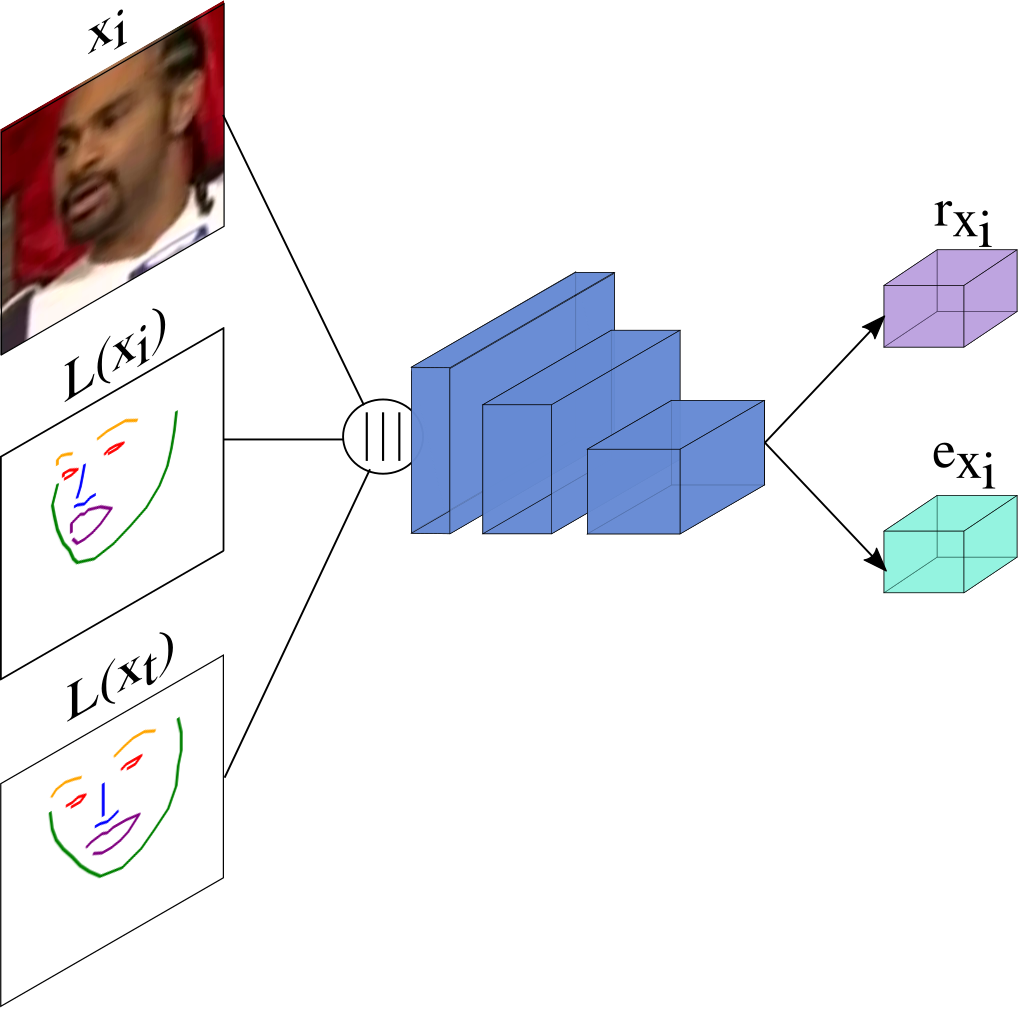}
        \caption{The structure of the Targeted Embedder. The real frame $x_i$, its landmark image $L(x_i)$ and target landmarks $L(x_t)$ are concatenated and processed to get $r_{x_i}$ and $e_{x_i}$.}
    \label{fig:temb}
\end{subfigure}
\caption[]{Embedder network}
\label{fig:embedder}
\end{figure}

The embedder computes a spatial embedding $e_{x_i}$ from a single input image: $e_{x_i} = E(x_i, L(x_i))$. Implementation details can be found in section \ref{sec:implementation}. In order to use multiple source images, a method to combine the embeddings must be devised. Zakharov et al. in \cite{zakharov2019fewshot} simply averages the one-dimensional tensors computed by the embedder. We observed that weighing the features by their relevance helps to better capture the identity, therefore a responsibility based combining method was developed (eq. \eqref{eq:1}, Figure \ref{fig:comb_emb}). To achieve this, the embedder will also output a weighing tensor $r_{x_i}$ representing the certainty for the computed features. The final identity embedding is calculated by a function $\Psi$ as:
\begin{equation}
    e_x = \Psi\left((e_{x_1}, r_{x_1}), \dots, (e_{x_K}, r_{x_K})\right) = \frac{\sum_{i \in \{i_1, \ldots, i_K\}}e_{x_i} \cdot r_{x_i}}{\sum_{i \in \{i_1, \ldots, i_K\}}r_{x_i}}
    \label{eq:1}
\end{equation}  
We explored the use of target landmarks $L(x_t)$ as input to the embedder along with $x_i$ and $L(x_i)$ and found this helpful for the generation process, since the identity features can be aligned to the final pose earlier. Hence, this is the version used in our experiments, denoted Targeted Embedder, as illustrated in Figure \ref{fig:temb}.

The generator starts from the target landmark and generates a new image $\hat{x}_t = G(L(x_t), e_x)$. The landmark image is provided through the input layer while the combined spatial embedding $e_x$ is used to modulate the activations at several resolutions with SPADE blocks.
In order to assess the quality of the generated image we employ two discriminators: identity discriminator and pose discriminator.

Identity discriminator $D_I(x_a, x_b)$ follows the multi-scale architecture from pix2pixHD \cite{wang2018pix2pixHD} and is used to estimate identity resemblance. At this stage we only consider characteristic traits and the results are expected to be invariant to pose. Thus, the input consists of two RGB images of the same person in arbitrary positions.

Pose discriminator $D_P(x_a, L(x_a))$ has the same architecture as $D_I$ with its own set of parameters. The network receives a frame and the appropriate target landmarks and checks the correspondence.  

While both discriminators will also assess general realism of a given image, $D_P$ is used specifically to avoid pose mismatch, whereas $D_I$ encourages identity preservation. This disengagement allows us to assign different importance factors to each discriminator while training. Different combinations may give results that correlate better with human visual perception.

\subsection{Training}
Videos are used as image sequences during training. For each epoch we sample a sequence of $K + 1$ distinct frames for each training video and compute landmark images: $L(x_{i_1}), L(x_{i_2}), ..., L(x_{i_{K+1}})$ using a pretrained face alignment network \cite{bulat2017far}. The embedding tensor $e_x$ is then computed according to (\ref{eq:1}). Using $L(x_t)$ and $e_x$, CainGAN generates the new image $\hat{x}_t$ which is further fed to the identity discriminator $D_I(\hat{x_t}, x_{i_1})$ along with a source frame $x_{i_1}$. The choice of the identity frame index $i_1$ is arbitrary as the generation process is not influenced by the order of input images. $\hat{x}_t$ and $L(x_t)$ are also fed to the pose discriminator $D_P(\hat{x_t}, L(x_t))$. Importance factors $\lambda_I$ and $\lambda_P$ control the weight of each discriminator in the objective function given by the hinge loss \cite{lim2017geometric,miyato2018spectral}:
\begin{equation}
    \begin{aligned}
    \mathcal{L}_{Adv}(G, E, D) = -\lambda_I \cdot D_I(G(L(x_t), e_x), x_{i_1}) \\ -\lambda_P \cdot D_P(G(L(x_t), e_x), L(x_t))
    \end{aligned}
    \label{eq:2}
\end{equation}

\begin{equation}
    \mathcal{L}_{Id}(D_I) = \max(0, 1 - D_I(x_t, x_{i_1})) + \max(0, 1 + D_I(\hat{x}_t, x_{i_1}))
    \label{eq:3}
\end{equation}
\begin{equation}
    \mathcal{L}_{Pose}(D_P) = \max(0, 1 - D_P(x_t, L(x_t)) + \max(0, 1 + D_P(\hat{x}_t, L(x_t))
    \label{eq:4}
\end{equation}

The identity and pose discriminators are updated using $\mathcal{L}_{Id}$ and $\mathcal{L}_{Pose}$, respectively. The weights of the generator and the embedder are updated together using the full objective:
\begin{equation}
    \begin{aligned}
    \mathcal{L}(G, E, D) = \mathcal{L}_{Adv}(G, E, D) +
    \lambda_{FM}(
    \lambda_{I} \cdot \mathcal{L}_{FM}(D_I) + \\ 
    \lambda_{P} \cdot \mathcal{L}_{FM}(D_P)) +
    \lambda_{VGG} \cdot \mathcal{L}_{VGG}
    \end{aligned}
    \label{eq:5}
\end{equation}

$\lambda_{FM}$ and $\lambda_{VGG}$ represent hyperparameters that control the importance of the loss in the full objective. $\mathcal{L}_{VGG}$ is a perceptual loss that compares features extracted at several layers by a pretrained VGGNet \cite{simonyan2014very} from the original and the generated image.
Feature matching loss $\mathcal{L}_{FM}$ is also a perceptual loss, comparing activations in the layers of the discriminator according to \eqref{eq:6}. Importance factors $\lambda_I$ and $\lambda_P$ also affect the weight of feature matching losses. The FM loss is similar to the one used in \cite{wang2018pix2pixHD} as we employ multiscale discriminators:
\begin{equation}
    \mathcal{L}_{FM}(D) = \sum_{i=1}^{S}\sum_{j=1}^{T}\frac{L_1(D_i^{(j)}(x_t, y), D_i^{(j)}(\hat{x}_t, y))}{N_j}
    \label{eq:6}
\end{equation}
where $S$ represents the number of scales, $T$ is the number of layers in $D$, $N_j$ is the number of elements in layer $j$ and $L_1$ denotes the standard Manhattan distance. $y$ is either the value of $L(x_t)$ when using the perceptual loss induced by the pose discriminator, or $x_{i_1}$ for the identity discriminator, respectively.

To ease the training process we start with a low importance factor for the identity discriminator and linearly increase it to its maximum value over the first 10 epochs. This allows the model to learn the easy task first, generating realistic face images in given pose, after which we gradually impose identity preservation.

We alternate between ($G$, $E$) and $D$ updates, with twice more steps for the discriminator and using two time-scale update rule \cite{heusel2017gans} to stabilize training.  

\subsection{Implementation Details}
\label{sec:implementation}
The generator resembles an encoder-decoder architecture for image translation. There are 4 downsamplings residual blocks with learned skip connections, 3 same resolution residual blocks and 4 upsampling layers. Instance normalization \cite{ulyanov2016instance} is used after every downsampling and upsampling layer. A SPADE residual block with spectral normalization \cite{miyato2018spectral,zhang2019self} is used after each upsampling. Nearest interpolation is used to bring the spatial embeddings to the appropriate resolution for each SPADE block. The discriminators are based on the architecture proposed by Wang et al. in \cite{wang2018pix2pixHD}, and use 2 scales as the images are relatively small. The embedder consists of 2 downsampling and 4 same resolution residual blocks which are shared while computing $r_{x_i}$ and ${e_{x_i}}$. Two independent same resolution residual blocks are then used to get $r_{x_i}$ and ${e_{x_i}}$.

\section{Experiments and Results}
\label{sec:experiments}

To evaluate our approach, we conduct extensive experiments on the VoxCeleb2 \cite{Chung18b} dataset. To emphasize the ability of our model to learn from less data, we only use a small subset of the actual dataset. Originally, the train set contains almost 6000 different speakers featuring more than a million videos. For our experiments we randomly selected 150 speakers and their corresponding videos (around 30,000), less than 3\% of the grand total. A video dataset was used since it is an accessible way to obtain multiple images with the same identity.

Quantitative comparison is performed against two baselines: pix2pixHD \cite{wang2018pix2pixHD} and previously state-of-the-art method for talking head generation \cite{zakharov2019fewshot} denoted FSHM (Few-shot Head Models). We trained the pix2pixHD model from scratch as described in the original paper and official implementation. In order to use the model without fine-tuning, the network input consists of all source frames and their landmarks as well as the target landmark; these are also given to the discriminator. We also implemented a version of FSHM (feed-forward only) in order to assess the results in a limited training data setting. 

Three different metrics are used to compare the described methods: structural similarity metric (SSIM) between the ground truth and the generated image is used to measure low-level structural similarity, cosine similarity (CSIM) between embedding vectors as computed by a pretrained face recognition network and Fr\'echet Inception Distance (FID) \cite{heusel2017gans} measuring perceptual realism which usually better captures the similarity of real and fake images. We follow the same training setup presented by Zakharov et al. \cite{zakharov2019fewshot}, using 50 video sequences with 32 test frames for each. 

\setlength{\tabcolsep}{1em}
\begin{table}[t]
\centering
\begin{tabular}{l c c c}
     Method (K) & SSIM $\uparrow$ & CSIM $\uparrow$ & FID $\downarrow$  \\
     \hline
     pix2pixHD(1) & 0.66 & 0.80 & 72.26 \\ 
     FSHM (1) & 0.64 & 0.72 & 93.17 \\
     FSHM-FF-full (1) & 0.61 & N/A & 46.61 \\
     FSHM-FT-full (1) & 0.64 & N/A & 48.5 \\
     CainGAN (1) & \textbf{0.69} & \textbf{0.85} & \textbf{35} \\
     \hline
     pix2pixHD(8) & 0.66 & 0.81 & 71.89 \\ 
     FSHM (8) & 0.65 & 0.73 & 83.13 \\
     FSHM-FF-full (8) & 0.64 & N/A & 42.2 \\
     FSHM-FT-full (8) & 0.68 & N/A & 42.2 \\
     CainGAN (8) & \textbf{0.77} & \textbf{0.91} & \textbf{24.92} \\
     \hline
\end{tabular}
\caption{K is the number of source frames used for testing. For SSIM and CSIM higher is better, for FID lower is better. CainGAN (8) was stopped after 20 epochs to avoid overfitting, all other models were trained for 30 epochs. The ``full'' suffix refers to the models being trained on the entire dataset. These results are taken from \cite{zakharov2019fewshot}. CSIM is not reported here, as a different face recognition network was used for the original results.
}
\label{tab:exp}
\end{table}

\begin{figure}[!t]
    \centering    
    \advance\leftskip-1.8cm
    \newlength\wid
    \setlength{\wid}{6.5em}
    \addtolength{\tabcolsep}{-5pt}
    \begin{tabular}{m{3em}m{\wid}m{\wid}m{\wid}m{\wid}m{\wid}}

        \centering{} &
        \multicolumn{1}{c}{\textbf{Source}} & 
        \multicolumn{1}{c}{\textbf{pix2pixHD}} & 
        \multicolumn{1}{c}{\textbf{FSHM}} & 
        \multicolumn{1}{c}{\textbf{CainGAN}} & 
        \multicolumn{1}{c}{\textbf{Real Image}} \\ 
        
        \centering{\textbf{K=1}}&
        \includegraphics[width=\wid]{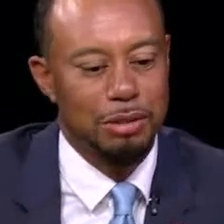} &
        \includegraphics[width=\wid]{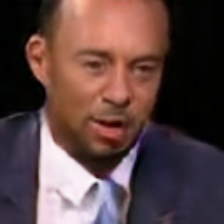}&
        \includegraphics[width=\wid]{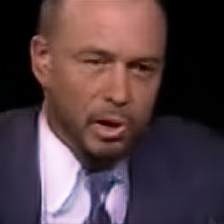}&
        \includegraphics[width=\wid]{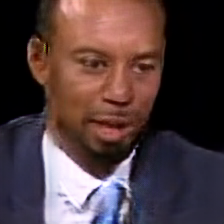}&
        \includegraphics[width=\wid]{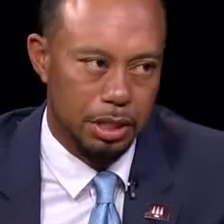}\\
        
        \centering{\textbf{K=8}}&
        \includegraphics[width=\wid]{Figures/d150/source.png}&
        \includegraphics[width=\wid]{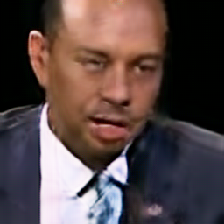}&
        \includegraphics[width=\wid]{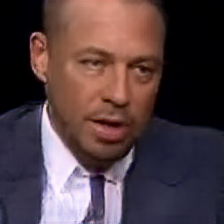}&
        \includegraphics[width=\wid]{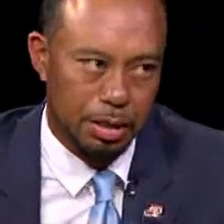}&
        \includegraphics[width=\wid]{Figures/d150/gt.png}\\
        
        \centering{\textbf{K=1}}&
        \includegraphics[width=\wid]{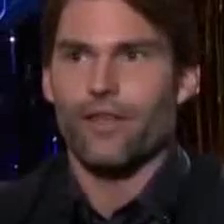} &
        \includegraphics[width=\wid]{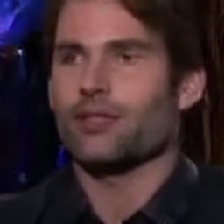}&
        \includegraphics[width=\wid]{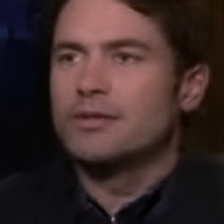}&
        \includegraphics[width=\wid]{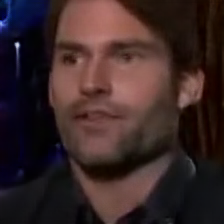}&
        \includegraphics[width=\wid]{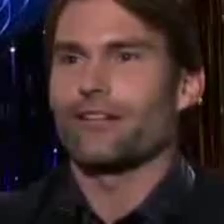}\\
        
        \centering{\textbf{K=8}}&
        \includegraphics[width=\wid]{Figures/d39/source.png}&
        \includegraphics[width=\wid]{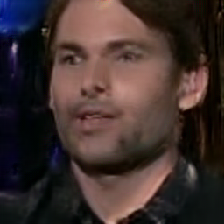}&
        \includegraphics[width=\wid]{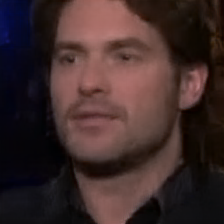}&
        \includegraphics[width=\wid]{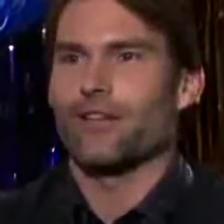}&
        \includegraphics[width=\wid]{Figures/d39/gt.png}\\
        
    \end{tabular}\vspace{-2pt}
    \caption{Visual assessment on the VoxCeleb2 dataset. First column represents the number of source frames, the next column illustrates one of the $K$ source images and the last column contains the ground truth ($x_t$) images. In between are the generated frames by different methods. The figure is best viewed in color.}
    \label{fig:qab}
\end{figure}

The comparison given in Table \ref{tab:exp} shows that CainGAN is able to get better quantitative results using only a fraction of the dataset. Additionally, the method is able to generate realistic images in the desired pose with a good preservation of identity. From qualitative comparison in Figure \ref{fig:qab} we can see that while FSHM can synthesize the face with the right alignment there is a high identity mismatch. Clearly, small amounts of training images severely affect the ability of the FSHM model to generalize to unseen faces. We also obtain the uncanny artifacts present in images generated by pix2pixHD, as reported in \cite{zakharov2019fewshot}.

\subsection{Ablation Study}
We performed an ablation study to analyze the influence of different components of our method. Quantitative results of the experiments are visible in Table \ref{tab:abl}. The variants are: CainGAN without targeting (CainGAN w/o T) where only the source frame and its landmarks are given to the embedder, CainGAN without discriminator importance weighing (CainGAN w/o I) where $\lambda_D = \lambda_I$ are fixed and CainGAN without responsibility based embedding mixing (CainGAN w/o R) where the weighted version in equation \eqref{eq:1} is replaced by:
\begin{equation}
    e_x = \frac{1}{K}\sum_{i \in \{i_1, \ldots, i_K\}} e_{x_i}
\end{equation}
This variant is not applicable for $K=1$ as in this case the two expressions yield the same result.

\setlength{\tabcolsep}{1em}
\begin{table}[!t]
\centering
\begin{tabular}{l c c c}
     Method (K) & SSIM $\uparrow$ & CSIM $\uparrow$ & FID $\downarrow$ \\
     \hline
     CainGAN w/o T (1) & \textbf{0.69} & \textbf{0.85} & 36.26 \\ 
     CainGAN w/o I (1) & 0.68 & 0.84 & 46.44 \\
     CainGAN (1) & \textbf{0.69} & \textbf{0.85} & \textbf{35} \\
     \hline
     CainGAN w/o T (8) & 0.72 & 0.87 & 38.08 \\ 
     CainGAN w/o I (8) & 0.76 & \textbf{0.91} & 28.72 \\
     CainGAN w/o R (8) & 0.75 & 0.90 & 30.05 \\
     CainGAN (8) & \textbf{0.77} & \textbf{0.91} & \textbf{24.92} \\
     \hline
\end{tabular}
\caption{Ablation study on selection of VoxCeleb2 dataset. All models were trained for 30 epochs, the best result between epochs 20 and 30 was reported. $K$ is the number of source frames.}
\label{tab:abl}
\end{table}

We can observe that all components are essential to obtain the best results. Using targeted embedder has a greater influence in the $K=8$ setting, which is expected since more images can benefit from early alignment.

\section{Conclusions and Future Work}
\label{sec:conclusion}
We introduced a new method for synthesizing images in novel poses while preserving the identity of a given subject. CainGAN  uses spatially adaptive normalization with a proper combining function of spatial feature maps in the embedding space. Experimental results show that CainGAN behaves better on scarce training sets compared to other methods. Furthermore, realistic images can be generated without the need for fine-tuning. The ablation study demonstrates CainGAN's non-redundant structure whereas the difference between scores in the $K=1$ and $K=8$ settings illustrate the ability to capitalize on more source images when available.
Further development directions include designing a method to extend the applicability of CainGAN beyond the task of self-reenactment and closing the gap between one-shot and multi-shot results by improving on single source image generation.

\begin{figure*}[h!]
\includegraphics[width=2.2cm]{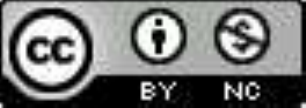}\\
Copyright $\copyright$2020 by the authors. Licensee Agora University, Oradea, Romania.\\
 This is an open access article distributed under the terms and conditions of the Creative Commons Attribution-NonCommercial 4.0 International License.\\
Journal's webpage: http://univagora.ro/jour/index.php/ijccc/ \\
\end{figure*}
\begin{figure*}[h!]
\centering
\includegraphics[width=3cm]{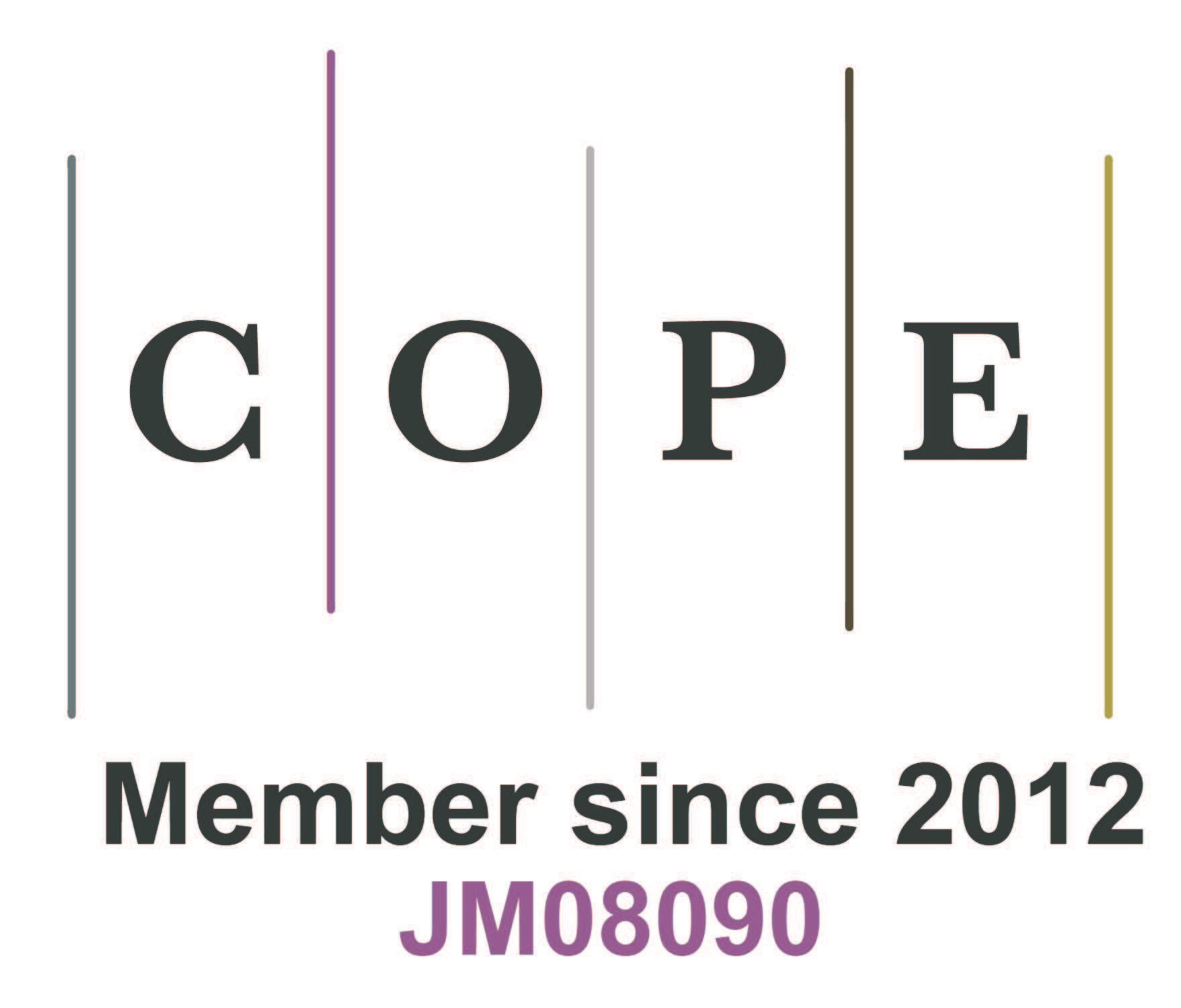}\\
This journal is a member of, and subscribes to the principles of,\\
 the Committee on Publication Ethics (COPE).\\
 https://publicationethics.org/members/international-journal-computers-communications-and-control
\end{figure*}

\label{LastPage}

\end{document}